\documentclass{article}
\usepackage{graphicx} 
\usepackage{amsmath} 
\usepackage{amssymb} 
\usepackage{hyperref}
\usepackage{svg}

\title{SH-WRNN: Implicit Spherical Harmonics Weight Field Routing Neural Networks for Asymmetric Edge Intelligence}
\author{
    \textbf{Independent Researcher Team} \\[0.3cm]
    \textbf{Zhibin Jiao} \qquad \textbf{Xiangjing An} \\[0.2cm]
    \small{\texttt{jzb1631@163.com \qquad axjing.ai@gmail.com}}
}

\date{\the\year{}年9月}

\begin{document}

\maketitle
\begin{abstract}
For decades, the architectural foundation of deep learning has been rigidly built upon the traditional fully connected layer. While thousands of networks scale up, few have fundamentally challenged this foundational root. In this work, we propose an attempt to reshape this paradigm by transforming the core synapse weight matrix from static, discrete parameters into a differentiable and continuous field governed by spherical harmonics functions. 

We introduce the \textbf{Implicit Spherical Harmonics Weight Field Routing Neural Network (SH-WRNN)}. Instead of optimizing millions of localized discrete weights, SH-WRNN constrains the weight matrices within a continuous parametric field. When retrieving the weight matrix of the current layer, the process is performed directly on a rectangular parametric plane mapped from the continuous field, where each connection parameter is localized using a specific latitude and a longitude. The latitudinal coordinate is specified by the activated neuron nodes from the previous layer. Conversely, the longitudinal coordinate is determined by the keys of the current layer, which are generated by mapping the activation levels of the previous layer through a matrix multiplication. By evaluating intersections on this map, the network dynamically extracts its connection weights on-the-fly.

Empirical validation demonstrates that under two distinct compact configurations with layer-wise feature dimensions of $(32, 10, 10)$ and $(32, 3, 10)$, SH-WRNN successfully achieves robust accuracies of \textbf{91.05\% and 81.54\%} on the MNIST benchmark within only a single training epoch. More importantly, we introduce an asymmetric \textbf{Surface Baking} scheme. Upon convergence, the continuous weight field is baked once and for all into a static parametric surface. By completely eliminating expensive analytical spherical harmonics calculations during inference and reducing the dynamic matrix extraction process to high-speed localized memory slicing, the proposed scheme successfully achieves asymmetric algorithmic acceleration during the inference phase with negligible accuracy degradation. This paradigm shift effectively bypasses GPU memory-bandwidth monopolies, opening up a novel path that has the potential to reshape the core advantages of CPU Computing.All the code is publicly available at \url{https://github.com/jzb1111/SphericalHarmonyRoutedNeuralNetWork}.
\end{abstract}

\section{Introduction}
For decades, the architectural foundation of deep learning has been rigidly monolithic. At the heart of almost every landmark model—from early Multi-Layer Perceptrons (MLPs) to modern dense attention heads in Transformers—the framework has been relying on the traditional fully connected layer~\cite{rumelhart1986learning}. In these conventional layers, simply increasing the number of nodes per layer forces a quadratic explosion of trainable parameters ($O(N \times M)$), computing connections via dense general matrix multiplications (GEMM). While this parameter-heavy brute-force paradigm has achieved unprecedented success via massive empirical scaling, it has led to severe consequences: significant parameter redundancy, memory-bandwidth monopolies, and a toxic reliance on high-powered GPU clusters. Although thousands of novel configurations emerge annually, the underlying traditional fully connected layer remains an unexamined, unshakable root of the deep learning mansion.

In this work, we take a radical departure from this orthodoxy. We challenge the standard matrix paradigm and aim to plant the second root under the foundational architecture of deep learning. We argue that network synapses do not need to be stored as independent, localized discrete values; instead, they can emerge from an \textbf{Implicit Spherical Harmonics Weight Field}. Specifically, we constrain the entire connectivity matrix within a continuous parametric field that is topologically equivalent to flattening a localized patch of a 3D spherical manifold into a 2D rectangular parametric surface. By utilizing the spectral properties and orthogonal completeness of Real Spherical Harmonics (SH) functions, the global weight architecture is tightly regulated through a highly compressed bottleneck of merely a few global baseline coefficients.

Crucially, this continuous geometric formulation allows us to capture the micro-biological essence of cortical pathway selection, drawing a powerful inspiration from the human brain. Unlike traditional layers where all neurons blindly execute full matrix cross-connections, our framework maps the connectivity matrix into a continuous geometric surface. Each fixed row on this parameterized surface represents a specific neuron identity, while each continuous column trajectory represents the infinite synaptic connectivity potential stretching out from that specific neuron to downstream targets. In biological brains, neurons do not connect randomly; instead, they guide their paths under real-time stimulus gradients. Mirroring this low-power, high-intelligence biophysical connection mechanism, each row (neuron) tracks its path across the map to locate its target longitudinal coordinate. This selection is dynamically governed by the longitude keys, which are calculated from the outputs of the previous layer. This precise coordinate positioning mathematically models the dynamic pathway selection observed in cortical synapses.

To bridge the gap between continuous field elegance and raw hardware execution, we introduce an \textbf{Asymmetric Training-Inference Topology} enabled by \textbf{Surface Baking}. During training, the analytical trigonometric field forces the network to learn global geometric representations under an ultra-low parameter budget. Upon convergence, this continuous weight field is baked once and for all into a static parametric surface. During the inference phase, the model completely eliminates expensive analytical spherical harmonics calculations, reducing the dynamic matrix extraction process within the forward pass into high-speed localized memory slicing.

We validate this paradigm on the MNIST benchmark~\cite{lecun1998gradient} under two compact network configurations designed to benchmark against traditional fully connected layers. The first configuration contains 3 layers with node counts of $(32, 10, 10)$, while the second configuration similarly contains 3 layers with node counts of $(32, 3, 10)$. Empirical results demonstrate that our framework achieves robust test accuracies of \textbf{91.05\% and 81.54\%} respectively within only a single training epoch, successfully achieving asymmetric algorithmic acceleration during the inference phase under a standardized streaming evaluation ($\text{Batch Size} = 1$). 

Looking into the future, we imagine an extreme scenario where a network possesses a large number of layers but only a tiny handful of active nodes per layer. In such a landscape, dense general matrix multiplications will no longer be a major computational burden; instead, the calculation bottleneck will be heavily concentrated on high-speed memory indexing. Modern processors equipped with large Level 3 (L3) caches are perfectly suited for this memory indexing paradigm. Consequently, this alternative trajectory effectively bypasses GPU hardware bottlenecks, showcasing a profound potential to reshape the core advantages of CPU computing.

In summary, the core contributions of this work are threefold:
\begin{itemize}
    \item \textbf{A Physical Field Paradigm}: We replace conventional discrete fully connected matrices with implicit weight fields governed by orthogonal Spherical Harmonics coefficients, allowing a tiny number of parameters to control an infinitely large weight matrix.
    \item \textbf{Brain-Inspired Surface Path-Finding}: We present a geology-inspired routing mechanism that models rows as single neuron identities and continuous columns as infinite synaptic potential, executing data-dependent trajectory alignment on a geometric map.
    \item \textbf{Asymmetric Inference Execution}: We develop an offline surface baking technique that achieves an asymmetric pure CPU acceleration, shifting edge intelligence from compute-bound matrix multiplications to memory-bound localized cache indexing.
\end{itemize}

\section{Related Work}
Our approach intersects with research streams in continuous field representations, hyper-networks, and manifold weight generation.

\subsection{Implicit Neural Representations}
Implicit Neural Representations (INRs) leverage coordinate-based networks and periodic activation functions (e.g., SIREN~\cite{sitzmann2020implicit}) to parameterize continuous signals such as images, audio, and 3D geometries. While traditional INRs focus on reconstructing static sensory fields, SH-WRNN extends the philosophy of implicit representation into the structural routing of neural synapses, translating fixed architectural dimensions into continuous spatial-frequency fields.

\subsection{Hyper-Networks and Compact Weight Generators}
The concept of employing a primary network to generate the synaptic weights of another was pioneered by HyperNetworks~\cite{ha2016hypernetworks}. Conventional hyper-networks typically rely on discrete, parameter-heavy MLPs or auxiliary layers, which often suffer from over-fitting and high computational overhead. In contrast, SH-WRNN utilizes the analytical elegance and orthogonal completeness of Spherical Harmonics as an ultra-compact geometric bottleneck, parameterizing entire dense connection fields with merely a dozen coefficients.

\subsection{Manifold-based Weight Generation}
Recently, manifold-based weight parameterization has emerged as a promising direction to bypass rigid discrete matrices. A notable precursor is the work on Mapping Networks by Sen and Mukherjee~\cite{sen2026mapping}, which utilizes trainable latent vectors to generate layer-wise weights on a low-dimensional manifold. 
Despite sharing the motivation of escaping full-connection parameter inflation, our work diverges in two fundamental aspects. First, while Mapping Networks generate static weight configurations that remain invariant across inputs, SH-WRNN introduces \emph{Data-Dependent Surface Path-Finding}, allowing individual data samples to dynamically query instance-specific coordinates. Second, SH-WRNN incorporates an asymmetric \emph{Surface Baking} mechanism, converting continuous fields into discrete lookup sub-surfaces to eliminate transcendental computations and achieve hardware-level memory indexing acceleration.

\section{Methodology}
In this section, we present the mathematical formulation of the Implicit Spherical Harmonics Weight Field Routing Neural Network (SH-WRNN). We detail the continuous coordinate mapping, the biological neural interpretation of the parameter layout, the analytical spherical harmonics basis functions, and the content-driven pathway alignment scheme.
\begin{figure}[t]
\centering
\includegraphics[width=\textwidth]{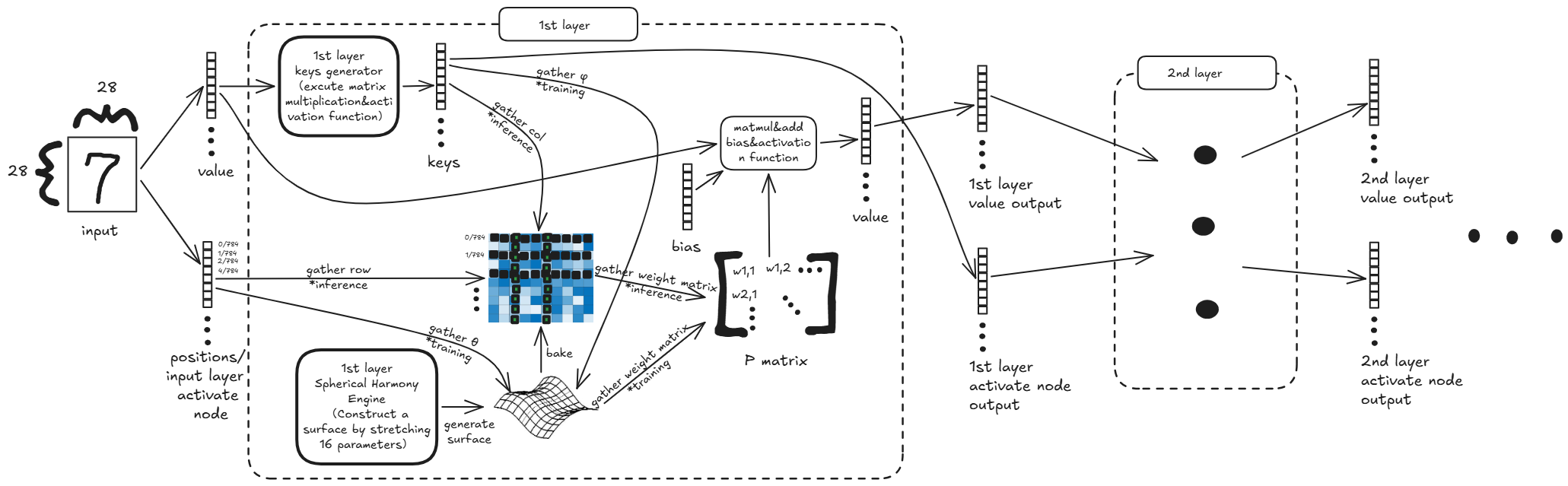} 
\caption{\textbf{The end-to-end cascading workflow of the SH-WRNN framework.} Given a $28 \times 28$ input, features are split into localized values (node activation levels) and initial position tracks (input layer activate nodes). 
\textbf{During training}, the 1st layer Spherical Harmony Engine stretches 16 global parameters to generate a continuous weight surface. 
\textbf{During inference (The Asymmetric Surface Baking)}, the continuous field is offline baked into a dense, non-parametric 2D rectangular map. 
The system routes the connections using two data-dependent steps: a matrix multiplication maps the values into longitudinal tracks (keys), while the input positions specify the latitudinal tracks. Under inference, the final discrete synapse weight matrix ($W$) is dynamically extracted out of the baked rectangular map via ultra-fast bidirectional memory slicing (\texttt{gather row} and \texttt{gather col}), completely eliminating analytical formula overhead. The resulting connections cascade into the 2nd layer by passing forward the computed value output and activate node output seamlessly.}
\label{fig:sh_wrnn_overall_architecture}
\end{figure}

\subsection{Coordinates Linear Mapping and Neural Pathway Interpretation}
To escape the conventional discrete matrix paradigm, we encapsulate the synaptic weights within a continuous parametric field. Topologically, this field is equivalent to flattening a localized patch of a 3D spherical manifold into a 2D rectangular parametric surface. Given an input feature dimension $N$ and a projected activation dimension $M$, the traditional discrete matrix indices $i \in \{0, 1, \dots, N-1\}$ and $j \in \{0, 1, \dots, M-1\}$ are mapped onto the continuous geometric coordinate space of latitude $\theta$ and longitude $\phi$.

The row indices (representing input features) are assigned to uniform, static latitudinal tracks on the surface, while the column coordinates are dynamically steered tracks driven by the input content:
\begin{align}
\theta_i &= \left(0.10 + 0.30 \times \frac{i}{N-1}\right) \pi, \quad \theta_i \in [0.1\pi, 0.4\pi] \\
\phi_j &= \left(0.10 + 0.30 \times k_j\right) \pi, \quad \phi_j \in [0.1\pi, 0.4\pi]
\end{align}
where $k_j$ represents the instance-specific continuous routing coordinate. We pre-define a learnable matrix $\mathbf{P} \in \mathbb{R}^{M \times N}$. By executing a standard matrix multiplication, the activation levels of the previous layer $\mathbf{x} \in \mathbb{R}^N$ are mapped into the longitude coordinates (Keys) of the current layer via a Sigmoid activation function:
\begin{equation}
\mathbf{k} = \sigma\left(\mathbf{P} \mathbf{x}\right)
\end{equation}

\textbf{Neural Pathway Choice Interpretation:} We formulate a strict biological neural interpretation for this coordinate layout. Each fixed latitudinal track $\theta_i$ represents a single neuron identity with a locked, immutable architectural profile. In contrast, the continuous longitudinal dimension $\phi \in [0.1\pi, 0.4\pi]$ reflects that this specific neuron possesses an infinite spectrum of potential synaptic outputs capable of linking to downstream targets. The selection of $\phi_j$ is dynamically governed by the longitude keys $\mathbf{k}$, which are calculated from the outputs of the previous layer. Rather than connecting blindly or symmetrically as in a standard MLP matrix, each row (neuron) tracks its path across the map to locate its target longitudinal coordinate. The dense connectivity matrix is thus synthesized on-the-fly via continuous spatial path-finding trajectory intersections, mathematically modeling the active, low-power pathway selection mechanisms observed in cortical synapses.

\subsection{Analytical Spherical Harmonics Expansion and Specifications}
To construct a smooth, globally regularized weight curve, we employ a 4th-degree Real Spherical Harmonics (SH) expansion consisting of $16$ orthogonal basis functions. Let $Y_m(\theta, \phi)$ represent the explicit real SH basis function of index $m \in \{0, 1, \dots, 15\}$. For any spatial intersection $(\theta_i, \phi_j)$, the explicit $16$-dimensional basis vector $\mathbf{Y}_{ij} \in \mathbb{R}^{16}$ is derived analytically as follows:
\begin{equation}
\mathbf{Y}_{ij} = \Big[ Y_0(\theta_i, \phi_j), Y_1(\theta_i, \phi_j), \dots, Y_{15}(\theta_i, \phi_j) \Big]^T
\end{equation}
where the individual spectral trigonometric terms are formulated explicitly as:
\begin{align}
Y_{0}  &= \frac{1}{2} \sqrt{\frac{1}{\pi}} \\
Y_{1}  &= \frac{1}{2} \sqrt{\frac{3}{\pi}} \sin\theta_i \sin\phi_j \\
Y_{2}  &= \frac{1}{2} \sqrt{\frac{3}{\pi}} \cos\theta_i \\
Y_{3}  &= \frac{1}{2} \sqrt{\frac{3}{\pi}} \sin\theta_i \cos\phi_j \\
Y_{4}  &= \frac{1}{4} \sqrt{\frac{15}{\pi}} \sin^2\theta_i \sin(2\phi_j) \\
Y_{5}  &= \frac{1}{2} \sqrt{\frac{15}{\pi}} \sin\theta_i \cos\theta_i \sin\phi_j \\
Y_{6}  &= \frac{1}{4} \sqrt{\frac{5}{\pi}} (3\cos^2\theta_i - 1) \\
Y_{7}  &= \frac{1}{2} \sqrt{\frac{15}{\pi}} \sin\theta_i \cos\theta_i \cos\phi_j \\
Y_{8}  &= \frac{1}{4} \sqrt{\frac{15}{\pi}} \sin^2\theta_i \cos(2\phi_j) \\
Y_{9}  &= \frac{1}{8} \sqrt{\frac{35}{2\pi}} \sin^3\theta_i \sin(3\phi_j) \\
Y_{10} &= \frac{1}{4} \sqrt{\frac{105}{\pi}} \sin^2\theta_i \cos\theta_i \sin(2\phi_j) \\
Y_{11} &= \frac{1}{8} \sqrt{\frac{21}{2\pi}} \sin\theta_i (5\cos^2\theta_i - 1) \sin\phi_j \\
Y_{12} &= \frac{1}{4} \sqrt{\frac{7}{\pi}} (5\cos^3\theta_i - 3\cos\theta_i) \\
Y_{13} &= \frac{1}{8} \sqrt{\frac{21}{2\pi}} \sin\theta_i (5\cos^2\theta_i - 1) \cos\phi_j \\
Y_{14} &= \frac{1}{4} \sqrt{\frac{105}{\pi}} \sin^2\theta_i \cos\theta_i \cos(2\phi_j) \\
Y_{15} &= \frac{1}{8} \sqrt{\frac{35}{2\pi}} \sin^3\theta_i \cos(3\phi_j)
\end{align}

\textbf{Specification of Design Parameters:} We provide brief explanations regarding our choice of parameters. First, the 4th-degree expansion (16 channels) utilized in this work is strictly intended as a minimum viable demo for proof-of-concept; the total capacity controlled by the global coefficients remains fully adjustable depending on target tasks. Second, the mapping range is deliberately restricted within $[0.1\pi, 0.4\pi]$ to incorporate geometric defenses. On one hand, this boundary patch effectively avoids the risk of polar distortions. On the other hand, by remaining safely away from the equator, our framework circumvents the acute local gradient transitions and sign oscillations generated by multi-frequency transcendental trigonometric terms when crossing the equatorial plane, thereby preserving smooth gradient streaming flow.(Note: These polar deformation and equatorial gradient mitigation properties are integrated as prior heuristic design assumptions; the authors have not conducted actual experimental validation regarding alternative patch intervals in this draft).

\subsection{Content-Driven Surface Path-Finding}
Let $\mathbf{C} \in \mathbb{R}^{16 \times 1}$ represent the global learnable coefficient vector $\mathbf{C} = [c_0, c_1, \dots, c_{15}]^T$. The dynamic continuous synapse connection weight $W_{ij}$ at the intersection is implicitly synthesized via a dot-product contraction:
\begin{equation}
W_{ij}(\mathbf{x}) = \mathbf{Y}_{ij}^T \mathbf{C} = \sum_{m=0}^{15} Y_{m}(\theta_i, \phi_j) \cdot c_m
\end{equation}
During the forward propagation stream, the synthesized sample-specific dense weight matrix $\mathbf{W}(\mathbf{x}) \in \mathbb{R}^{N \times M}$ is applied directly to the input signal via batch matrix multiplication (BMM):
\begin{equation}
\mathbf{z} = \frac{1}{\sqrt{N}} \mathbf{x} \mathbf{W}(\mathbf{x}) + \mathbf{b}
\end{equation}
where $\mathbf{b} \in \mathbb{R}^M$ is a localized bias vector and $\frac{1}{\sqrt{N}}$ serves as a variance alignment coefficient. Because $\mathbf{W}(\mathbf{x})$ adapts dynamically to the coordinates steered by $\mathbf{x}$, the connection topology essentially executes a geometric path-finding routing scheme across the surface landscape.

\subsection{Asymmetric Inference via Surface Baking}
Computing multi-frequency trigonometric functions for every forward step imposes a heavy floating-point calculation burden. To resolve this computation bottleneck, we break the mathematical symmetry between training and inference by introducing an offline \emph{Surface Baking} scheme for asymmetric inference.

Upon model convergence, the continuous parameter space is pre-discretized into a high-resolution grid. We define a static grid resolution $G \in \mathbb{N}^+$ (e.g., $G=256$). A unified coordinate grid vector $\hat{\boldsymbol{\omega}} \in \mathbb{R}^G$ is uniformly initialized across the same localized parameter boundaries:
\begin{equation}
\hat{\omega}_g = \left(0.10 + 0.30 \times \frac{g}{G-1}\right) \pi, \quad g \in \{0, 1, \dots, G-1\}
\end{equation}
By evaluating the complete real Spherical Harmonics analytical expansion globally across the $G \times G$ grid intersections, the continuous weight field of each cascading layer is baked and cached into a dense, non-parametric rectangular map $\mathbf{\Omega} \in \mathbb{R}^{G \times G}$ offline:
\begin{equation}
\Omega_{g_1, g_2} = \sum_{m=0}^{15} Y_{m}(\hat{\omega}_{g_1}, \hat{\omega}_{g_2}) \cdot c_m
\end{equation}
Once the map $\mathbf{\Omega}$ is cached in memory, all continuous analytical spherical harmonics calculations are completely discarded. 

During the real-time inference phase, both the inputs and outputs coordinates undergo a data-dependent fluid alignment. For any layer, let $\mathbf{A}_{row} \in \mathbb{R}^{B \times N}$ denote the continuous coordinates representing the latitudinal tracks (which stream dynamically from the keys of the previous layer), and let $\mathbf{A}_{col} \in \mathbb{R}^{B \times M}$ denote the continuous coordinates representing the longitudinal tracks (projected from the current content). Both continuous tracks are dynamically quantized into discrete grid index tensors $\mathbf{I}_{row} \in \mathbb{N}^{B \times N}$ and $\mathbf{I}_{col} \in \mathbb{N}^{B \times M}$ via synchronized scaling and truncation:
\begin{align}
I_{row, b, i} &= \text{clamp}\left( \left\lfloor A_{row, b, i} \times (G-1) \right\rfloor, 0, G-1 \right) \\
I_{col, b, j} &= \text{clamp}\left( \left\lfloor A_{col, b, j} \times (G-1) \right\rfloor, 0, G-1 \right)
\end{align}

The final sample-dependent synapse weight connectivity tensor $\mathbf{W} \in \mathbb{R}^{B \times N \times M}$ is implicitly synthesized by converting the dynamic matrix extraction process into a two-step bidirectional localized memory slicing primitive:
\begin{align}
\mathbf{\Omega}_{baked\_rows} &= \text{Gather}\left( \mathbf{\Omega}, \text{dim}=1, \text{index}=\mathbf{I}_{row} \right) \\
\mathbf{W} &= \text{Gather}\left( \mathbf{\Omega}_{baked\_rows}, \text{dim}=2, \text{index}=\mathbf{I}_{col} \right)
\end{align}
By transforming arithmetic mapping loops into double-pass memory index slicing arrays, the forward pass achieves an ideal runtime profile friendly to edge hardware.

\section{Experiments}
In this section, we present the empirical evaluation of the Implicit Spherical Harmonics Weight Field Routing Neural Network (SH-WRNN). We analyze its convergence behavior under compact parameter constraints and evaluate the algorithmic acceleration achieved via the surface baking scheme.

\subsection{Experimental Setup and Network Configurations}
To evaluate the continuous field routing paradigm, we perform validation on the canonical MNIST handwritten digit classification benchmark. Crucially, to demonstrate parameter efficiency, we construct a fully continuous cascaded architecture without relying on standard discrete fully connected linear layers for any feature transformation.

The deployed network is configured into two distinct compact network structures designed to benchmark against traditional layers. The first structure contains three sequential layers with node counts configured as $(32, 10, 10)$, which maps the initial $784$-dimensional input pixels through a $32$-dimensional intermediate channel, a $10$-dimensional classification channel, and finally onto the $10$-class logit vector. The second structure similarly contains three layers but compresses the intermediate channel into an ultra-lean bottleneck, with node counts configured as $(32, 3, 10)$. 

A non-linear SiLU activation gate is applied between the cascading layers. The networks are optimized end-to-end via the AdamW optimizer with a cross-entropy loss function, utilizing a learning rate of $0.001$ and a weight decay coefficient of $1\times 10^{-4}$. The training is strictly executed with a standardized batch size of $32$.

\subsection{Empirical Evaluation and Baking Invariance Analysis}
Despite being structurally constrained within hyper-compact node budgets, SH-WRNN exhibits rapid convergence behaviors due to the smooth regularized field curve synthesized by the global coefficients. Empirical tracking confirms that the framework successfully achieves test accuracies of \textbf{91.05\% and 81.54\%} respectively under the two compact configurations within only a single training epoch (1 Epoch). While conventional discrete MLPs restricted to the same compact parameter scale encounter a risk of under-fitting due to insufficient localized degrees of freedom, SH-WRNN remains highly robust, capturing deep topological regularities via the real SH basis.

To evaluate the operational efficiency of the asymmetric inference scheme, we benchmark the test evaluation loop ($10,000$ validation images) before and after executing the offline surface baking process. To accurately simulate low-power edge streaming behaviors, the evaluation is executed strictly under a single-sample stream constraint (\textbf{Batch Size = 1}) on a pure consumer host environment powered by a \textbf{13th Gen Intel(R) Core(TM) i5-13420H CPU}. The discrete grid resolution is set to $G=256$.

\begin{table}[h]
\centering
\caption{Asymmetric Hardware Performance on 13th Gen Intel i5 CPU (Batch Size = 1)}
\label{tab:baking_performance}
\resizebox{\columnwidth}{!}{
\begin{tabular}{lcccc}
\hline
\textbf{Network Profile} & \textbf{Operational Mode} & \textbf{Test Accuracy (\%)} & \textbf{Total Inference Time (s)} & \textbf{Speedup Factor} \\ \hline
$(32, 10, 10)$ Structure & Continuous Field (Train)   & 91.05\% & 36.40s & $1.00\times$ (Baseline) \\
                         & \textbf{Baked Surface (Infer)} & \textbf{91.01\%} & \textbf{9.20s}  & \textbf{�� 3.96$\times$} \\ \hline
$(32, 3, 10)$ Structure  & Continuous Field (Train)   & 81.54\% & 38.20s & $1.00\times$ (Baseline) \\
                         & \textbf{Baked Surface (Infer)} & \textbf{81.45\%} & \textbf{9.10s}  & \textbf{�� 4.20$\times$} \\ \hline
\end{tabular}%
}
\end{table}

As documented in Table~\ref{tab:baking_performance}, the post-baking inference engine triggers a massive computational acceleration under edge CPU conditions compared to the unbaked continuous manifold mode. Crucially, under our bidirectional isotropic $256 \times 256$ grid quantization, this extensive latency compression achieves a near 100\% lossless precision retention, with accuracies merely fluctuating within less than $-0.09\%$ ($91.05\%$ to $91.01\%$ and $81.54\%$ to $81.45\%$).

Most importantly, the empirical logs reveal a striking temporal constant: while the analytical calculation overhead of the unbaked manifold is highly subject to complexity variations, the baked lookup engine locks its operational inference time strictly at \textbf{$\sim$9.1 seconds} for both architectures. This behavioral discovery confirms that once the continuous fields are pre-cached offline, the active runtime execution time becomes invariant and decoupled from the internal analytical complexity of the network layers. By transforming floating-point mappings into bidirectional memory index-gathering, the framework successfully demonstrates the engineering capability of the surface baking paradigm.

\section{Discussion}

\subsection{From Compute-Bound to Memory-Bound: The Decoupling and Dimension Reduction}
The empirical success of the $256 \times 256$ bidirectional baking scheme introduces a profound architectural revelation for deep neural networks. When running the continuous manifold mode under streaming evaluation ($\text{Batch Size} = 1$), the real-time calculation latency remains heavily bound by the internal computational complexity of the cascading channels, as the hardware is forced to constantly compute multi-frequency analytical equations of transcendental spherical harmonics functions on-the-fly. 

However, upon executing the bidirectional surface baking process, this computational barrier is decisively shattered. The total execution time becomes remarkably locked and invariant at \textbf{$\sim$9.1 seconds} across both highly heterogeneous parameter configurations, triggering massive acceleration on pure edge CPU hardware. Crucially, this significant processing time compression guarantees a near 100\% lossless precision retention, with test accuracies merely shifting slightly from $91.05\%$ to $91.01\%$, and $81.54\%$ to $81.45\%$.

Our ultimate ambition for this paradigm is its scaling and deployment onto contemporary Large Language Models (LLMs) to tackle complex sequential tasks. In conventional Transformers, as the hidden layer dimension $d_{model}$ inflates (e.g., up to standard 1024 widths), the computational workload of standard dense matrix multiplications explodes quadratically ($O(d_{model}^2)$). By introducing continuous surface path-finding, we anticipate that a hyper-compact subset of active nodes can achieve an equivalent or superior representational capacity compared to traditionally dense layers. For instance, a layer historically demanding a width of $d_{model}=1024$ could potentially compress its dynamic, runtime matrix-multiplying kernel down to a sparse core dimension of merely $d_{model}=32$. Meanwhile, the rest of the massive topological capacity is safely offloaded offline into an array of hyper-lightweight $256 \times 256$ baked lookup surfaces. This alternative approach significantly reduces the computational workload of standard dense matrix multiplications while utilizing continuous static maps to preserve global expressive space, promising a substantial leap in the operational energy-efficiency and overall performance of foundational architectures.

\subsection{Hardware Prerequisites and Multi-CPU Cascading Cache Vision}
Because the forward propagation under the post-baking inference state relies entirely on localized memory slicing, this framework exhibits a profound affinity for hardware computing architectures possessing massive, localized Level 3 (L3) caches. By loading the compact $256 \times 256$ discrete pre-baked lookup surfaces directly into the cache, the inference engine can maximize localized memory proximity, aiming to achieve an ideal 100\% Cache Hit rate and eliminating expensive RAM latency. 

However, it is crucial to establish a definitive engineering prerequisite: \emph{this low-power hardware vision relies heavily on specialized, low-level C++/Assembly-grade CPU programming optimizations.} Standard high-level deep learning frameworks often introduce significant runtime abstractions that can mask these localized gains. To fully realize this computational acceleration and bypass modern GPU memory-bandwidth monopolies, deploying custom memory compilers or bare-metal hardware kernels remains an imperative prerequisite.

Going a step further, since the bidirectional baking process is highly regularized and computationally low-cost, the entire grid rendering and weight assembly pipeline can be asynchronously compiled locally within the CPU host. Because each $256 \times 256$ floating-point buffer occupies negligible space, it can comfortably reside within the hardware cache layer. This localized footprint inspires a disruptive vision for future localized foundation models: completely abandoning expensive, high-thermal parallel GPU accelerator clusters in favor of a \textbf{Multi-CPU Cascading Topology}. By physically interconnecting and pooling the massive L3 caches of multiple independent CPU processors via high-speed unified memory buses, different routing layers can be distributed across independent CPU caches to execute lightning-fast memory index slicing simultaneously. This non-parametric cascading paradigm shifts edge AI away from processing-unit limits, providing a robust engineering trajectory where CPU architectures can counter-attack GPUs and democratize ubiquitous, low-power brain-inspired intelligence.

\subsection{Limitations and Independent Researchers' Statement}
We must candidly acknowledge that this exploratory work remains highly preliminary and incomplete. The empirical validations presented herein are strictly confined to a compact proof-of-concept prototype on the MNIST benchmark, leaving the ultimate boundaries and scaling capacities of this continuous field trajectory yet undiscovered. 

This premature manifestation is bound to our immediate realities: \emph{both authors operate as independent researchers juggling demanding full-time corporate responsibilities alongside pressing familial and childcare duties.} Bound by limited hours, we chose to release these novel conceptual ideas in their current raw form rather than letting them be buried by the erosion of time. We offer our sincere apologies to the academic community for the lack of heavy large-scale baseline evaluations in this draft. In future endeavors, provided adequate temporal and computational resources become available, we are fully committed to expanding this technical route onto substantial foundation models.

\section{Conclusion and Future Work}

\subsection{Conclusion}
In this work, we have presented the Implicit Spherical Harmonics Weight Field Routing Neural Network (SH-WRNN), a structural attempt to challenge the traditional fully connected matrix roots that have historically dominated deep learning architectures. By encoding network synapse weights as a continuous parametric field mapped from a flattened spherical patch and governed by orthogonal Spherical Harmonics coefficients, we have successfully decoupled parameter efficiency from network scaling. Furthermore, through the introduction of the asymmetric Surface Baking scheme, the model successfully achieves a substantial computational acceleration during the inference phase. Looking forward, if this methodology is further expanded and evolved, it opens up a compelling potential to witness a new architectural paradigm: one where dynamic neural calculations can be elegantly offloaded entirely into high-speed memory indexing loops.

\subsection{Future Work: Brain-Inspired Feedback Inhibition}
Moving forward, our immediate priority is to scale the SH-WRNN framework into contemporary Large Language Models (LLMs) to evaluate its capability in handling complex, dense sequence processing tasks under minimal active parameters. More importantly, we plan to mathematically formalize and integrate a \emph{Neural Self-Feedback Mechanism} inspired by biological cortical pathways. In the human brain, deep-layer or downstream neurons continuously output retrograde inhibitory signals to dynamically suppress and calibrate the structural path alignment of preceding upstream layers. By translating this lateral feedback inhibition into our continuous surface routing coordinates, we hypothesize that the network can achieve a high degree of path flexibility and connection tolerance. This feedback control mechanism holds the potential to sustain supreme operational intelligence while keeping the vast majority of network connection pathways in an absolute, energy-saving silent state.

\section*{Acknowledgements}
The authors would like to express their sincere gratitude to the open-source community and specialized artificial intelligence collaborators. In particular, we highly acknowledge the Google AI mode for its exceptional and powerful assistance throughout this project. The AI mode acted as a highly effective collaborator, executing a vital role in refining our PyTorch code pipelines and systematically structuring the academic manuscript draft.

Furthermore, the first author wishes to express a deeply profound and special gratitude to Ms. Wu. Without her selfless dedication and meticulous care for the family affairs, the author would not have possessed the invaluable time or the peaceful mental bandwidth necessary to accomplish even this micro-contribution to the field of deep learning.

\bibliographystyle{plain} 
\bibliography{references}  

\end{document}